\documentclass[10pt,twocolumn,letterpaper]{article}

\PassOptionsToPackage{table}{xcolor}

\usepackage[pagenumbers]{iccv}

\definecolor{iccvblue}{rgb}{0.21,0.49,0.74}

\usepackage{amsmath} 
\usepackage[utf8]{inputenc}
\usepackage[T1]{fontenc}
\usepackage[pagebackref,breaklinks,colorlinks,allcolors=iccvblue]{hyperref}
\usepackage{url}
\usepackage{booktabs}
\usepackage{amsfonts}
\usepackage{nicefrac}
\usepackage{microtype}
\usepackage{multirow}
\usepackage{graphicx}
\usepackage[ruled]{algorithm2e}
\usepackage{xspace}

\def\paperID{4516}
\def\confName{ICCV}
\def\confYear{2025}

\newcommand{\methodname}{ERNet\xspace}

\definecolor{colorfirst}{rgb}{.866,.945, 0.831}
\definecolor{colorsecond}{rgb}{1, 0.98, 0.83}
\definecolor{colorthird}{rgb}{0.76, 0.87, 0.92}

\newcommand{\cellfirst}{\cellcolor{colorfirst}}
\newcommand{\cellsecond}{\cellcolor{colorsecond}}

\newcommand{\Warp}{\mathbf{\mathcal{W}}_{i}}

\title{\methodname: Efficient Non-Rigid Registration Network for Point Sequences}

\author{
Guangzhao He\qquad
Yuxi Xiao\qquad
Zhen Xu\qquad
Xiaowei Zhou\qquad
Sida Peng$^{\text{†}}$\\[0.8em]
Zhejiang University
}

\begin{document}

\twocolumn[{
    \maketitle
    \begin{center}
        \captionsetup{type=figure}
        \vspace{-5mm}
        \includegraphics[width=\textwidth]{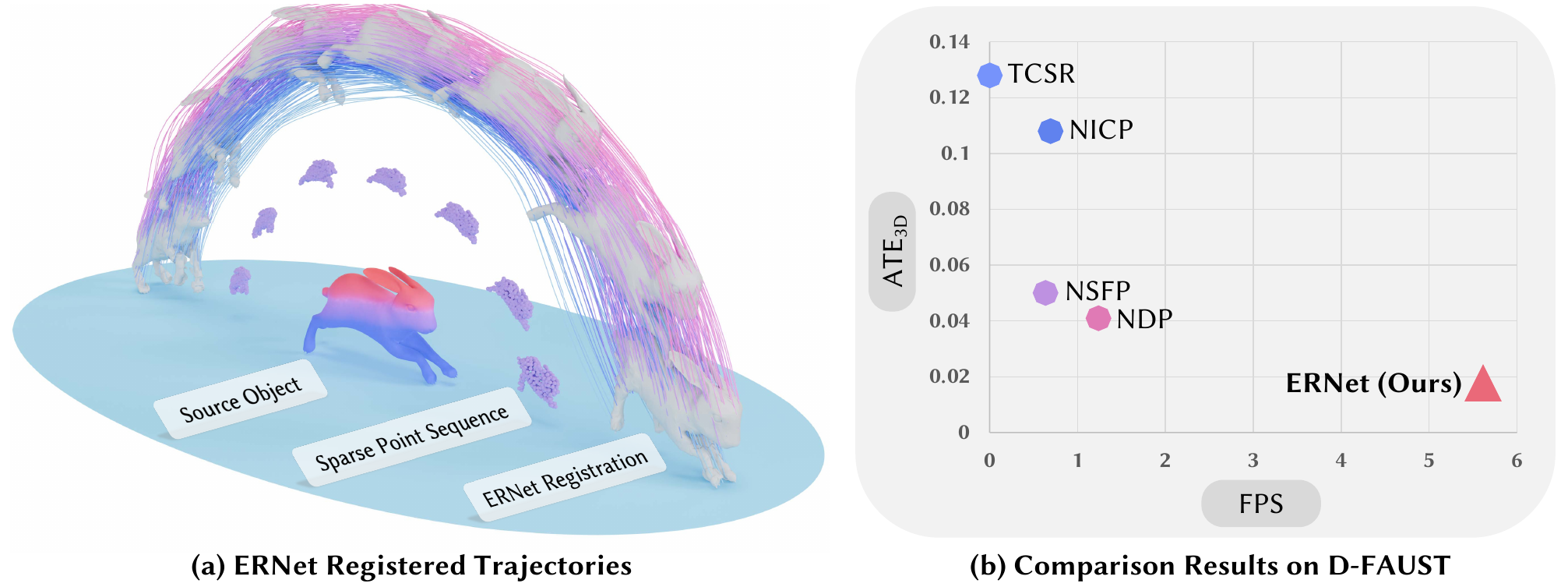}
        \vspace{-5mm}
        \caption{ 
            (a) Given a source object (center bunny) and a sequence of sparse or partial point clouds (purple bunnies), \methodname efficiently and accurately predicts feed-forward registrations. 
            The gradient texture on the center bunny corresponds to the colors of its vertex-wise trajectories shown above.
            (b) \methodname achieves the lowest Average Trajectory Error in 3D ($ATE_{3D}$) among all baselines on the D-FAUST dataset while delivering over 4x speedup in FPS.
        }
        \vspace{0mm}
    \end{center}
}]

\let\thefootnote\relax\footnotetext{The authors are affiliated with the State Key Lab of CAD\&CG. $^{\text{†}}$Corresponding author.}

\begin{abstract}

Registering an object shape to a sequence of point clouds undergoing non-rigid deformation is a long-standing challenge. 
The key difficulties stem from two factors: (i) the presence of local minima due to the non-convexity of registration objectives, especially under noisy or partial inputs, which hinders accurate and robust deformation estimation, and (ii) error accumulation over long sequences, leading to tracking failures.
To address these challenges, we introduce to adopt a scalable data-driven approach and propose \methodname, an efficient feed-forward model trained on large deformation datasets.
It is designed to handle noisy and partial inputs while effectively leveraging temporal information for accurate and consistent sequential registration.
The key to our design is predicting a sequence of deformation graphs through a two-stage pipeline, which first estimates frame-wise coarse graph nodes for robust initialization, before refining their trajectories over time in a sliding-window fashion.
Extensive experiments show that our proposed approach (i) outperforms previous state-of-the-art on both the DeformingThings4D and D-FAUST datasets, and (ii) achieves more than 4x speedup compared to the previous best, offering significant efficiency improvement.
\end{abstract}
    
\section{Introduction}
\label{sec:intro}
As Carl Jung once said, “In all chaos there is a cosmos, in all disorder a secret order.”
Predicting structured motion representation, more specifically 3D trajectories, within unordered point sequences has always been a fundamental challenge in computer vision and robotics, with wide applications among dynamic reconstruction, scene understanding, robot manipulation and more.
In this work, we target on the problem of sequential non-rigid registration, which aims to register a source mesh of an object onto a sequence of observed point clouds, often sparse or partial, resulting in dense trajectories of the object over time.
Traditional methods \cite{thinplate-splines, global-nonrigid-alignment} generally formulate this task as an optimization problem that solves point-wise transformations by minimizing the distance between source and target point clouds. 
While many follow-up works \cite{DBLP:journals/cgf/ChangZ09, DBLP:journals/cgf/LiSP08, DBLP:journals/tog/SumnerSP07,
DBLP:journals/tog/SharfALGSAC08,
allen2003space} have introduced regularizers, such as As-Rigid-As-Possible constraint \cite{DBLP:journals/tog/SharfALGSAC08,
DBLP:conf/cvpr/NewcombeFS15} and deformation graph \cite{DBLP:journals/tog/TevsBWIBKS12,allen2003space}, to improve the robustness of their optimization process, they still easily get stuck in local optima due to the non-convexity of the objective functions.

With the success of deep learning and neural networks, researchers have been exploring the potentials of neural networks in solving this problem. One type of this research leverages the functionality of neural network in representing deformations~\cite{m2vs, oflow, cadex, lpdc, deformation_graph} to register point clouds by optimizing neural deformation fields, partially alleviating the local optimal problem. 
However, because of their optimization-based nature, they still struggle to generalize to noisy or partial inputs. To address this, another line of work \cite{nsfp, lepard, geotr, roitr} propose to train registration prior models, leveraging learned deformation knowledge to improve robustness and mitigate local minima. While effective, these methods perform frame-wise registration rather than directly operating on sequential data, leading to time-consuming processing and issues such as error accumulation and temporal inconsistency.

In this paper, we propose a novel framework, named \methodname, for efficient sequential non-rigid registration.
Our core innovation lies in a feed-forward network that predicts a sequence of deformation graphs based on spatio-temporal matching across point clouds.
Specifically, instead of implicitly predicting a set of 3D shape keypoints as the graph nodes~\cite{jakab2021keypointdeformer}, we explicitly obtain a set of nodes from the source object using the farthest point sampling algorithm, considering its applicability for diverse 3D shapes.
Then, our approach leverages a coarse-to-fine strategy to regress frame-wise node positions throughout the point cloud sequence, which first estimates coarse node positions in each frame through spatial matching between source nodes and point clouds, and then globally refines the node trajectories across frames in a sliding-window manner.
As shown in our experiments, the use of deformation graphs enables efficient feed-forward registration, while the two-stage strategy, taking into account the spatio-temporal relationship between node positions, enhances robustness and consistency under noisy and partial input.

A remaining problem is how to obtain the blending weights and SE(3) transformations of the graph nodes to drive the source object onto each frame.
We find that naively predicting these properties with a neural network suffers from their non-linearity and high-dimensional characteristics, resulting in sub-optimal performance.
To overcome this issue, our approach exploits the local rigidity of non-rigidly deforming objects to infer these properties.
During the refinement stage, we jointly predict the radii of nodes, which can be easily inferred based on local geometric cues and motion correlations.
Then, using the radial basis function, we define blending weights for each pair of source point and deformation node.
For the SE(3) transformation, we follow Procrustes analysis and group a set of local nodes to calculate their rotations and translations through the singular value decomposition algorithm.
Our experimental results demonstrate that this strategy works well and produces high-quality registration results.

We evaluate our approach on the DeformingThings4D~\cite{li20214dcomplete} and D-FAUST~\cite{bogo2017dfaust} datasets, which are challenging benchmarks for estimating sequential non-rigid deformation.
Across these datasets, our method achieves state-of-the-art performance in both accuracy and efficiency.
Additionally, we demonstrate its robustness to sparse and partial inputs, and conduct ablation studies to validate the effectiveness of our proposed modules.

In summary, our contributions are:
\begin{itemize}
\item We propose an architecture for feed-forward sequential non-rigid registration, incorporating a novel two-stage prediction strategy for improved robustness and temporal consistency.
\item We propose to directly regress deformation graphs as an efficient representation for non-rigid registration. 
\item We evaluate the proposed pipeline on several different deformation datasets, and demonstrate significant improvements in both accuracy and speed compared to the state-of-the-art.
\end{itemize}

\section{Related Work}

\paragraph{Representations of Deformation Field.}
Representations of deformation field often involve the trade-off between its expressiveness and the computational cost. 
Point-wise affine transformation is one of the simplest ways to define the deformation field~\cite{pointwise_field, huang2011global, allen2003space}. 
Although it is highly expressive for modeling complex motion, its redundancy in degrees of freedom often leads to expensive computational cost and under-constrained problem. 
To mitigate this, deformation graphs~\cite{deformGraph} are proposed to represent point-wise motion with a set of graph nodes, each associated with an SE(3) transformation.
Individual deformation for each point can then be calculated with weighted skinning.
Since the number of graph nodes is typically several orders of magnitude smaller than the original point cloud, they can offer significant efficiency improvement while reducing the solution space for deformation optimization.
Recently, with the success of implicit neural fields~\cite{mildenhall2020nerf, occnet}, some works~\cite{cadex, Nerfies, NSFF} propose to represent the deformation field as a continuous implicit mapping from 3D coordinates to deformation vectors, often implemented as a multilayer perceptron (MLP). 
Furthermore, to reduce the high complexity of modeling deformations with neural networks, \cite{ndp} proposes to use several levels of MLPs to hierarchically represent deformations with different levels of details. 
However, such implicit representations are often inefficient, requiring per-frame optimization which is impractical for registering long sequences.
In contrast, we adopt deformation graphs as an explicit representation of deformation, and propose to regress them with a neural network in a feed-forward manner, striking a balance between efficiency and expressiveness.

\paragraph{Non-rigid Registration.} 
Non-rigid registration aims to find point-wise deformation from the source to the target point cloud. 
Registration algorithms are typically designed to accommodate specific deformation representations. 
Non-rigid iterative closest point (NICP)~\cite{NICP} is a classic optimization-based registration algorithm, and is commonly used for solving either point-wise transformations or deformation graphs. 
In order to regularize the optimization process and better preserve local topology details, As-Rigid-As-Possible~\cite{huang2021arapreg} constraint is proposed to regularize the neighboring nodes to deform as rigidly as possible. 
However, they are still sensitive to initialization and often get stuck in local minima. 
To further regularize the deformation and incorporate temporal information, OccupancyFlow~\cite{oflow} first proposes to model sequential deformation as a neural velocity field and predicts deformation with an ordinary differentiable equation (ODE) solver. 
Similarly, CaDeX~\cite{cadex} employs implicit neural network by formulating the deformation field as bijective mappings between each frame and a shared canonical space.
Nevertheless, both methods struggle to capture high-frequency deformations effectively with implicit networks.
Another line of work~\cite{reg_network, Deepgraph, flownet3d} predicts dense point correspondences, and estimates frame-wise deformations in a feed-forward manner.
While these methods achieve impressive accuracy, extending them for efficient sequential registration is non-trivial due to their limited ability to aggregate temporal information.
Moreover, their computational complexity scale poorly with the number of input points, making them inefficient for large-scale inputs.
In this paper, we address these limitations by making efficient feed-forward prediction of sparse deformation graphs, and incorporating a coarse-to-fine strategy to fully utilize temporal information for robust and accurate registration.

\paragraph{Temporal Tracking.}
Recent advancements in Tracking Any Points (TAP) algorithms~\cite{cotracker, doersch2022tap, doersch2023tapir, spatialtracker} have established an effective framework for tracking arbitrary 2D points over long video sequences. 
More specifically, CoTracker~\cite{cotracker} introduces a sliding-window approach with overlapping segments, ensuring temporal consistency through a spatio-temporal transformer. 
SpatialTracker~\cite{spatialtracker} extends this concept by lifting 2D image features using off-the-shelf depth estimators and perform tracking in the 3D space.
However, these methods struggle to re-track occluded points over extended sequences due to limited window size.
To address this limitation, we propose a two-stage registration pipeline.
In the first stage, we employ a dedicated frame-wise matching module for coarse yet robust initialization, effectively handling occlusions and preventing error accumulation in long sequences.
In the second stage, we incorporate a 3D temporal refinement module for sequential non-rigid registration, which predicts temporally consistent and accurate trajectories for graph nodes while also estimating node radii for weighted skinning.

\section{Method}

\label{method}
\begin{figure*}[t]
    \centering
    \includegraphics[width=1.0\textwidth]{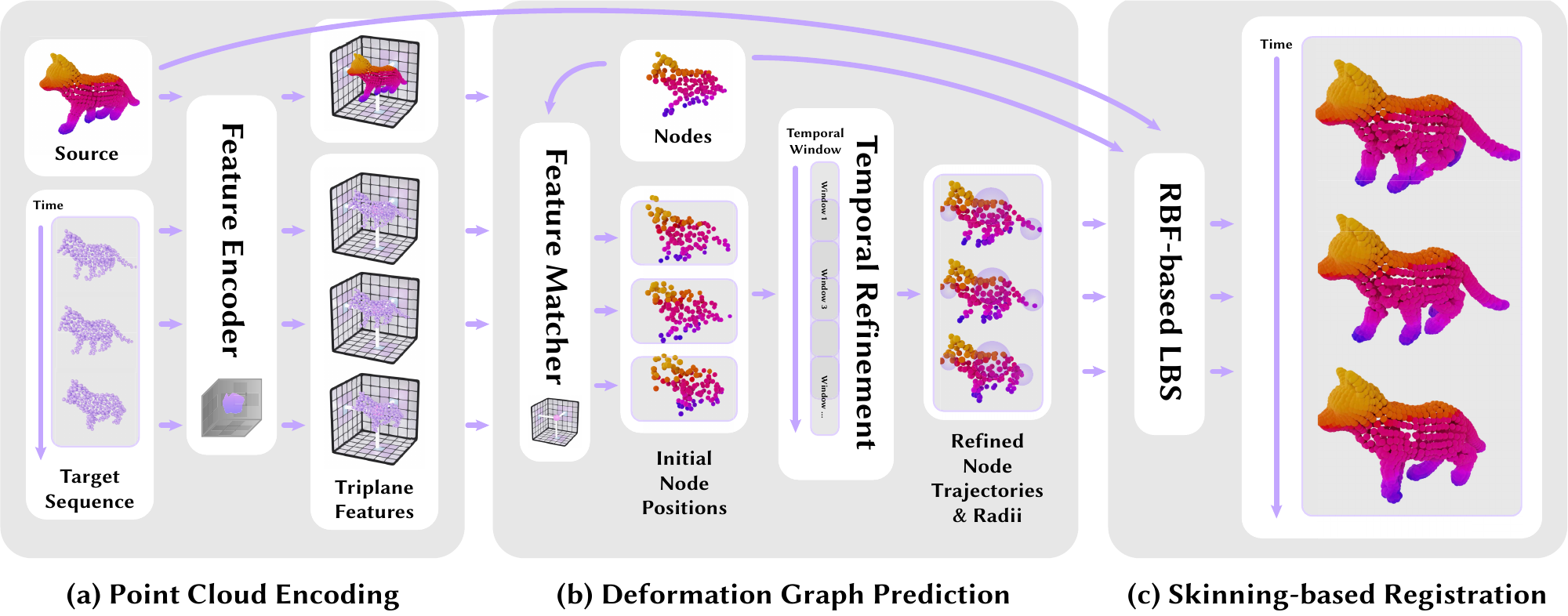}
    \vspace{-17pt}
    \caption{
        \textbf{Overview of our proposed pipeline.}
        (a) Given a source point cloud and input target point cloud sequence, we first encode them independently using a shared local feature encoder and splat per-point features onto triplane grids (Sec.~\ref{method:feature}).
        (b) Then we initialize graph nodes based on source point cloud and perform the coarse-to-fine matching to predict the node positions and radii of the deformation graph using encoded features (Sec.~\ref{method:transformer}).
        (c) With the predicted node trajectories and radii, we calculate node transformations via the Procrustes analysis and utilize the RBF-based LBS to produce dense registration (Sec.~\ref{method:skinning}).
    }
    \label{fig:pipeline}
    \vspace{-10pt}
\end{figure*}

Given a dense source point cloud $\mathbf{X}_s\in\mathbb{R}^{N_s\times3}$ and a temporal sequence of sparse target points $\mathcal{P}=\{\mathbf{P}_i\in\mathbb{R}^{N_i\times3}\,|\,i=1,\cdots,T\}$, our goal is to design and train a feed-forward model to predict a series of non-rigid deformation fields: 
\begin{equation}
    \Warp:(\mathbf{X}_s,\mathcal{P})\mapsto\mathbf{X}_p^i\,,\,i=1,\cdots,T,
\end{equation}
which register the source point cloud $\mathbf{X}_s$ to its corresponding point cloud $\mathbf{X}_p^i$ at each frame. 
In this paper, we focus on developing an learning-based framework that can generalize to diverse object categories and shape deformations while producing high-accuracy and temporally consistent registrations throughout the point cloud sequence.  

The overview of our approach is illustrated in Fig.~\ref{fig:pipeline}. 
We first encode the source point cloud $\mathbf{X}_s$ and target point cloud sequence $\mathcal{P}$ with an efficient triplane encoder (Sec.~\ref{method:feature}). 
Then we represent the non-rigid deformations as a sequence of graph nodes. 
We initialize their per-frame positions with a matching network, and apply a spatio-temporal transformer to iteratively refine their trajectories and radii (Sec.~\ref{method:transformer}). 
Finally, dense deformation fields are applied to $\mathbf{X}_s$ to complete the sequential registration using radial basis blend skinning (Sec.~\ref{method:skinning}).
To train our model, we introduce a two-stage training strategy that first pre-trains the matching module before optimizing the remaining pipeline end-to-end (Sec.~\ref{method:training}).

\subsection{Efficient point cloud encoding}
\label{method:feature}
We encode point clouds using a triplane encoder, which encodes arbitrary number of 3D points into three orthogonal feature planes. 
With this design, per-point features are compressed continuously into three planes, enabling fast indexing and reducing memory consumption while preserving detailed geometry information.
Both per-frame point sequence $\mathcal{P}$ and source point cloud $\mathbf{X}_s$ are encoded using the same encoder with shared weights.
The overall point cloud encoder $\mathcal{E}$ is then formalized as: 
\begin{equation}
    \mathcal{F}_{xy},\mathcal{F}_{yz},\mathcal{F}_{xz}=\mathcal{E}(\mathbf{X})\,,\,\mathbf{X}\in\{\mathbf{X}_s\}\cup\mathcal{P}.
\end{equation}

\paragraph{Local feature encoder.}
To extract low-level geometry information, we utilize a modified shallow PointNet \citep{pointnet} with local pooling layers instead of global pooling. 
More specifically, each residual block is followed by a local pooling operation, which we achieve by projecting per-point features orthographically onto triplane grids and aggregating the averaged results as local features.
The aggregated features are then concatenated with per-point features to form the inputs for following residual blocks.
This ensures that the encoded feature contains semi-global information while maintaining accuracy around local areas, which benefits accurate feature matching.

\paragraph{Triplane feature maps.} 
After embedding local feature for every point, the geometry is still discrete and contains spatial gaps.
One naive solution would be to linearly interpolate between nearby points to form a continuous feature field. 
However, this leads to inaccurate approximation of geometry features and introduces expensive nearest-neighbors searching overhead.
Similar to Xiao et al.~\citep{spatialtracker}, we project per-point features onto triplane grids to enable fast indexing, and apply a shallow U-Net \citep{unet} to fill gaps and complete the geometry features. 
Compared to using 3D CNN, our approach of triplane factorization significantly reduces memory consumption.

\subsection{Deformation graph prediction}
\label{method:transformer}
\paragraph{Deformation graph representation.}
Deformation graph is an efficient and global representation for motion~\cite{deformGraph}. It consists of a set of sparse graph nodes $\mathcal{G} = (\mathcal{V}, \mathcal{R}, \mathcal{T})$, and each graph node is associated with
its positions $\mathcal{V} = \{\mathbf{V}_p\in\mathbb{R}^{3}\,|\,p=1,\cdots,B\}$, radii $\mathcal{R} = \{\mathbf{R}_p\in\mathbb{R}\,|\,p=1,\cdots,B\}$ and series of $\mathbf{SE}(3)$ transformations $\mathcal{T}=\{\mathcal{T}^{i}_{p}\,|\,i=1,\cdots,T\,,p=1,\cdots,B\}$ resulting the final deformation. Here, we sample $B$ points from the source point cloud as positions of graph nodes using the farthest point sampling algorithm.

\paragraph{Challenges for deformation graph prediction.}
For deformation graph estimation, one of the naive solution is to extract the features of graph nodes at each time step, and regress the graph nodes attributes directly. However, we found that primitively regressing graph nodes via a neural network is very challenging, especially the $\mathbf{SE}(3)$ transformations. Meanwhile, the temporal consistency is hard to be ensured for the cases of large motions. To overcome this, we firstly estimate the trajectories of graph nodes following coarse-to-fine fashion. The trajectories are actually the linear offsets which is easy for network prediction. The coarse-to-fine strategy first coarsely match the graph nodes with each frame and refine them with an iterative spatio-temporal transformer, which ensures the temporal and global consistency. After that, we leverage the local rigidity to estimate $\mathbf{SE}(3)$ transformations from the predicted trajectories by Procrutes analysis.

\paragraph{Node-to-frame matching.}
Since $\mathbf{X}_s$ and $\mathcal{P}$ can be in arbitrary poses with large motions, we find it necessary to first initialize graph nodes positions for each frame by using a node-to-frame matching network $\mathcal{\phi}$. It takes the source nodes $\mathbf{V}_s$ as well as their embedded triplane features $\mathbf{F}_s$, and produces a coarse registration for all given frames:
\begin{equation}
    \mathbf{V}_{p}^i=\mathcal{\phi}(\mathbf{V}_s,\mathbf{F}_s,(\mathcal{F}_{xy}^i,\mathcal{F}_{yz}^i,\mathcal{F}_{xz}^i)).
\end{equation}
More specifically, $\mathcal{\phi}$ is a spatial transformer module, and its input token in frame $i$ is defined as:
\begin{equation}
    \label{eq:node-to-frame_input}
    \mathbf{G}^i=(\gamma(\mathbf{V}_p^i),\gamma(\mathbf{V}_p^i-\mathbf{V}_s),\mathbf{F}_p^i,\mathbf{F}_s,\mathbf{C}_p^i),
\end{equation}
where $\gamma$ denotes sinusoidal positional encoding function,
$\mathbf{V}_p^i$ and $\mathbf{F}_p^i$ are node positions and features initialized with $\mathbf{V}_s$ and $\mathbf{F}_s$, 
and $\mathbf{C}_p^i$ denotes the correlations between node features and triplane features near $\mathbf{V}_p^i$.
The transformer consists of multiple self-attention blocks, and outputs residuals for updating node positions and features.
We apply the transformer for $\mathbf{M}=6$ times, and after the $m$-th iteration the updated results are given by:
\begin{equation}
    \label{eq:node-to-frame_update}
    (\mathbf{V}_p^m,\mathbf{F}_p^m)=(\mathbf{V}_p^{m-1},\mathbf{F}_p^{m-1})+\mathcal{\phi}(\mathbf{G}^{m-1}).
\end{equation}
Here the sequence index $i$ is omitted for clarity.

\paragraph{Spatio-temporal refinement.}
After applying node-to-frame matching, the initialized node positions are still inaccurate and lack temporal consistency.
To solve this issue, we utilize a spatio-temporal transformer $\Phi$ to refine 3D node trajectories based on triplane geometry features.
The $T$ frames are first partitioned into overlapping windows of the same length $T_w$, where the second half of frame overlaps with the first half in the following window. 
Temporal node graph updates are carried out one window at a time, so that long sequences can be handled in an online fashion.
When updating node positions within a window $W=[i_0,i_0+T_w)$, the transformer $\Phi$'s input token is defined as:
\begin{equation}
    \mathbf{G}=\cup_{i=i_0}^{i_0+T_w}\{(\gamma(\mathbf{V}_p^i),\gamma(\mathbf{V}_p^i-\mathbf{V}_p^{i_0}),\mathbf{F}_p^i,\mathbf{F}_s,\mathbf{C}_p^i)\,\},
\end{equation}
where notations are similar to those in Eq.~\ref{eq:node-to-frame_input} except that $\mathbf{V}_p^i$ is initialized using node-to-frame matching results, and $\mathbf{V}_p^{i_0}$ denotes node positions for the first frame in the window.
Similar to Eq.~\ref{eq:node-to-frame_update}, the output from the transformer $\Phi$ are used to to update the node positions and features.
Here we perform trajectory updates within one window for $M=6$ iterations, and move on to the next window by initializing its first $\frac{T_w}{2}$ frames with the results of last $\frac{T_w}{2}$ frames in the previous window.

\paragraph{Transformation estimation.}
After recovering the temporal consistent nodes trajectories, our approach uses the predicted node trajectories to estimate transformations from source object to a particular time step for each graph node.
We first find a set of source graph nodes that exhibits the local rigidity, then find the corresponding graph nodes at the target time step, and finally solve the transformations between two sets of nodes via the Procrustes analysis process.

Specifically, for a source graph node $\mathbf{v}_s$, our approach first searches for its $K$-nearest nodes $\mathcal{N}=\{\mathbf{v}_s^k\,|\,k=1,\cdots,K\}$ in source graph nodes $\mathbf{V}_s$ as the candidates, and finds the trajectories for all $K$ nodes $\{\mathbf{v}_i^k\,|\,i=1,\cdots,T\,,\,k=1,\cdots,K\}$.
We assume that the nearest neighbor node $\mathbf{v}_s^1$ is always correctly assigned, and include other graph nodes if they satisfy the following criteria:
\begin{equation}
    \label{eq:node_assignment}
\mathcal{N}_f'=
    \{\mathbf{v}_s^k\,|\,
    \max_i \{|\frac{\|\mathbf{v}_i^k-\mathbf{v}_i^0\|_2}{\|\mathbf{v}_s^k-\mathbf{v}_s^0\|_2}-1|\}<\epsilon 
    \,,\,k=2,\cdots,K\},
\end{equation}
where $\epsilon$ is initialized as 0.2.
This is used to only include nodes that stay relatively rigid w.r.t. $\mathbf{v}_s$. 
The resulting node set is thus denoted as $\mathcal{N}_f=\{\mathbf{v}_s^1\}\cup\mathcal{N}_f'$.

To enhance the robustness of the Procrustes analysis, it is important that the number of nodes within $\mathcal{N}_{f}$ is not less than 4.
Therefore, if the number of nodes obtained is less than 4, we increment $\epsilon$ by 0.1 and recalculate Eq.~\ref{eq:node_assignment} to derive a new set of nodes.
Then, our approach regards these nodes as one rigid part, and uses the estimated node trajectories to produce corresponding graph nodes at the target time step.
Finally, the transformation from two sets of graph nodes are calculated via the Procrustes analysis process.
The detailed calculations are provided in the supplemental material.

\paragraph{Node radius prediction.}
To convert the deformation graph into a dense deformation field, the linear blend skinning (LBS) algorithm~\cite{DBLP:journals/tog/SumnerSP07} is applied here to construct the warping function. However, direct estimate the blending weights is non trivial due to its high dimensionality and non-linearity. Therefore, we utilize an extra spatio-temporal transformer $\Phi_R$ to regress a radius $\mathbf{R}_p$ for each node. Then the estimated radii are used to calculate the blending weights with Radial Base Functions (RBF), which is introduced in Sec.~\ref{method:skinning}.

\subsection{Skinning-based registration}
\label{method:skinning}

\paragraph{Selective graph nodes assignment.}
Skinning is an indispensable step for converting the estimated deformation graph into the dense warping function, which is used to establish correspondences between source and observed point clouds.
The key challenges here lie on correct nodes assignment for each source point. The incorrect neighbours selection will lead to the wrong skinned deformation when the topology changes occur. 

To increase the robustness to close-to-open topology changes, we propose to assign graph nodes to each point by considering the local rigidity.
For each point $\mathbf{x}_s$ in source point cloud $\mathbf{X}_s$, we firstly search for its $K'_{init}$-nearest nodes in source graph nodes $\mathbf{V}_s$ as the candidates, and obtain corresponding nodes based on the estimated node trajectories.
Then, these nodes are filtered based on a process similar to Eq.~\ref{eq:node_assignment}.
Here $\epsilon$ is set as 0.2.
This metric effectively filter those graph nodes with substantial relative position shifts, which always indicates for the close-open topology change.

\begin{table*}[t]
    \caption{\textbf{Quantitative results} on the test subsets of DT4D-H and D-FAUST datasets.
    Note that, our approach is only trained on D-FAUST and DT4D-A datasets and can generalize to DT4D-H dataset.
    Green and yellow cell colors indicate the best and the second best results, respectively.
    }
    \label{tab:seq-reg}
    \centering
    \resizebox{0.9\textwidth}{!}{
    \begin{tabular}{lcccccccc}
      \toprule
      \multirow{2}{*}{Method} &             
      \multicolumn{4}{c}{DT4D-H} &
      \multicolumn{4}{c}{D-FAUST} \\
      \cmidrule(lr){2-5} \cmidrule(lr){6-9}
                              & $ATE_{3D}\downarrow$ & $\delta_{0.01}\uparrow$ & $\delta_{0.05}\uparrow$ & $T_{avg}\downarrow$ & $ATE_{3D}\downarrow$ & $\delta_{0.01}\uparrow$ & $\delta_{0.05}\uparrow$ & $T_{avg}\downarrow$ \\
      \midrule
      C-NICP~\cite{NICP} & 0.107                & 0.038                   & 0.167                   & 1.569               & 0.108                & 0.022                   & 0.188                   & 1.445               \\
      C-NSFP~\cite{nsfp} & 0.070                & 0.062               & 0.418                   & 2.040               & 0.050                & 0.056                   & 0.437                   & 1.570               \\
      C-NDP~\cite{ndp}   & \cellsecond 0.058                & \cellsecond 0.192                   & \cellsecond 0.603                   & \cellsecond 0.832               & \cellsecond 0.041                & \cellsecond 0.057                   & \cellsecond 0.522                   & \cellsecond 0.808               \\
      \methodname (ours)       & \cellfirst{0.037}       & \cellfirst{0.238}          & \cellfirst{0.678}          & \cellfirst{0.188}      & \cellfirst{0.016}       & \cellfirst{0.313}          & \cellfirst{0.863}          & \cellfirst{0.178}      \\
    \bottomrule
    \end{tabular}
    }
\end{table*}

\paragraph{Radial basis skinning.}
After knowing the graph node neighbors for each source point, the deformation for each point can be calculated with the LBS algorithm.
In order to simplify the blending process and ensure the training stable, we select the radial basis function (RBF) to calculate the blended weights for skinning.
Specifically, for a point $\mathbf{x}_s$, we calculate the blending weight of its assigned node $\mathbf{v}_s(\mathbf{x}_s)$ with respect to frame $i$ using:
\begin{equation}
    \mathbf{w}_i=\exp(-\frac{\|\mathbf{x}_s-\mathbf{v}_s(\mathbf{x}_s)\|_2^2}{2\,{\mathbf{r}_i(\mathbf{x}_s)}^2}).
\end{equation}
where $\mathbf{r}_i(\mathbf{x}_s)$ is the node radius estimated in Sec.~\ref{method:transformer}.
To deform the point $\mathbf{x}_s$ from the source object to the position $\mathbf{x}_i$ at frame $i$, our approach first finds its assigned graph nodes $\{\mathbf{v}_s^k(\mathbf{x}_s)\}$ and corresponding transformations $\{\mathcal{T}_i^k(\mathbf{v}_s)\}$, and then performs the LBS algorithm:
\begin{equation}
\mathbf{x}_i=\frac{\Sigma_{k=1}^{K'}\mathbf{w}_i^k\mathcal{T}_i^k(\mathbf{v}_s)}{\Sigma_{k=1}^K\mathbf{w}_i^k},
\end{equation}
where $K'$ is the number of graph nodes assigned to point $\mathbf{x}_s$.
This is performed for every point in $\mathbf{X}_s$ across all frames, forming the final sequential registrations $\{\mathbf{X}_p^i\,|\,i=1,\cdots,T\}$.

\subsection{Training}
\label{method:training}

Our model parameters include triplane geometry encoder $\mathcal{E}$, node-to-frame matching network $\phi$ and spatio-temporal transformers $\Phi_R$ and $\Phi$.
To optimize them, we design an efficient and stable two-stage training strategy.

We first start with training $\mathcal{E}$ and $\phi$ from scratch on point cloud pairs, where $\mathcal{P}$ only contains one target point cloud.
The optimization is regularized using node position regression loss:
\begin{equation}
    \mathcal{L}_{match}=\Sigma_{m=1}^M\alpha^{M-m}\|\hat{\mathbf{V}}_p^m-\mathbf{V}_p\|_1,
\end{equation}
where $\hat{\mathbf{V}}_p^m$ denotes the predicted node position at $m$-th iteration, and $\mathbf{V}_p$ denotes the ground-truth position.
We set $\alpha=0.8$ to guide the matching network to gradually update nodes from source position towards target position.

After the encoder $\mathcal{E}$ and matching network $\phi$ converges, we freeze their weights to go on and optimize transformers $\Phi$ and $\Phi_{R}$.
This enables us to train them on larger windows and longer sequences with wider temporal context. We give the same supervision on the predictions of each sliding window. So, to avoid the complicated notations, we introduce the total loss on arbitrary sliding window $T_w$. The total loss is a combination of registration loss, node regression loss and local rigidity constraint:
\begin{equation}
\mathcal{L}_{\text{total}}= \Sigma_{m=1}^{M}\alpha^{M-m}(\mathcal{L}_{\text{reg}}^{m}+\lambda_{\text{node}}\mathcal{L}_{\text{node}}^{m}+\lambda_{\text{rigid}}\mathcal{L}_{\text{rigid}}^{m}),
\end{equation}
where $\lambda_{\text{node}}$ and $\lambda_{\text{rigid}}$ are weight coefficients, which in practice is set to $1.0$ and $0.1$, respectively.
The losses are summed for every frame in every iteration across all windows.
We present the details of these losses in the supplementary material.

\section{Experiments}

\subsection{Implementation details}
\label{experiments}
Our model is trained from scratch with two 48GB A6000 GPUs. 
The first stage training takes 200k iterations (7 days) to converge, while the second stage is trained for another 300k iterations.
For triplane feature encoding, we utilize five consecutive encoding-splatting blocks and set the plane resolution to be $256\times256$ with $128$ feature channels.
For node graph estimation, we set $B=256$ and find it to be the balance between modeling complex deformations and computational efficiency, which is shown in our ablation study~\ref{ablation:num-node}.
In the second training stage, we set the window size $T_w$ to $8$ and the total training frames to be $12$.
Both node-to-frame transformer $\phi$ and sequential transformer $\Phi$ have a layer depth of $12$, while the depth of node radius regressor $\Phi_{R}$ is set to $6$ for efficiency.
When performing blend skinning for source points, we choose $K=4$ to allow smooth transformation while maintaining accuracy by enforcing spatial locality.

\begin{figure*}[t]
    \centering
    \includegraphics[width=1.0\textwidth]{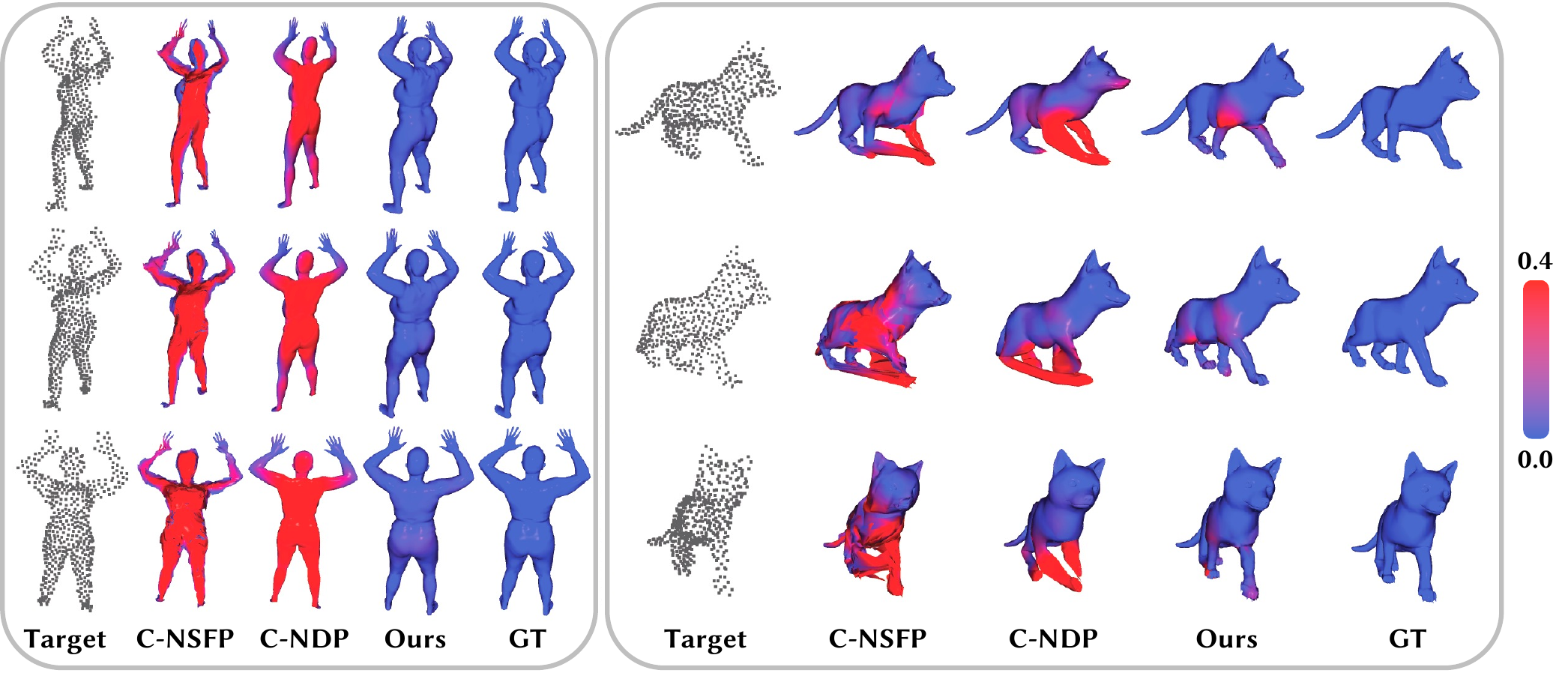}
    \vspace{-17pt}
    \caption{
        \textbf{Qualitative results} on two challenging examples from the depth-sampled D-FAUST and DT4D-A datasets.
        The point color reflects the L2 distance from ground truth, where blue indicates less error and red indicates more.
    }
    \label{fig:quality}
    \vspace{-5pt}
\end{figure*}

\subsection{Datasets and metrics}
\label{exp:dataset}
To showcase the generalizability and accuracy of our design, we train one single model on two challenging deformation datasets: DeformingThings4D~\cite{li20214dcomplete} and Dynamic FAUST~\cite{bogo2017dfaust}.

\paragraph{DeformingThings4D.}
It is a synthetic dataset consisted of 1772 animal deformation sequences (DT4D-A) and 200 human deformation sequences (DT4D-H).
It is especially challenging thanks to its wide variety of complex shapes and inclusion of highly-deformed sequences.
Following CaDeX~\cite{cadex}, we filter out the sequences where meshes contain ill-behaved areas and partition a subset of the DT4D-A dataset into training (75\%), validation (7.5\%) and test (17.5\%) subsets.
We do not include the DT4D-H dataset for training and sample test subset for evaluation of generalizability.

\paragraph{Dynamic FAUST (D-FAUST).} 
It is a human motion dataset consisting of 10 subjects and 129 deformation sequences. 
It contains challenging large deformations, e.g., "running on spot" and "punching".
To produce partial point clouds, we render depth images through a randomly posed camera per sequence and back-projecting them into 3D space.
We follow \cite{oflow} and partition it into training (70\%), validation (10\%), and test (20\%) subsets.

\paragraph{Metrics.}
To evaluate sequential registration accuracy, we calculate the average trajectory error $ATE_{3D}$ for each scene, which is the average l1 distance between the predicted registration targets and ground truths.
In addition, we compute $\delta_{0.01}$ and $\delta_{0.05}$ to evaluate the stability of registration accuracy, where $\delta_{0.01}$ denotes the fraction of predicted points that are within 0.01 unit length from ground truths, with $\delta_{0.05}$ using a relaxed threshold at 0.05 unit length.
To evaluate method efficiency, we provide average registration time per frame $T_{avg}$ in second for every method.

\subsection{Sequential non-rigid registration}

Since there are no existing works that can be directly used for sequential registration, we construct three baselines using pair-wise non-rigid registration methods, namely NDP~\cite{ndp}, NICP~\cite{NICP} and NSFP~\cite{nsfp}.
To perform registration, we first utilize these methods to predict the deformation from source point cloud to the first frame, and then chaining the registrations by iteratively predicting deformations based on the last predicted registration.
We refer to these baselines as chained NDP (C-NDP), chained NICP (C-NICP) and chained NSFP (C-NSFP) respectively.
We evaluate our method as well as the constructed baselines using the test sets of DT4D-H and D-FAUST.

The results are shown in Tab.~\ref{tab:seq-reg}.
Our method outperforms all baselines by a large margin both in accuracy and efficiency, even though all baselines are optimized per sequence while we utilize one single model across all data.
This is especially surprising since we do not utilize the DT4D-H dataset for training.
The results demonstrate that our model is highly generalizable and scalable thanks to our trajectory representation of node graphs.
We show qualitative results comparing our method with baseline methods in Fig.~\ref{fig:quality}

\subsection{Ablation study and analysis}

\begin{figure*}[t]
    \centering
    \includegraphics[width=1.0\textwidth]{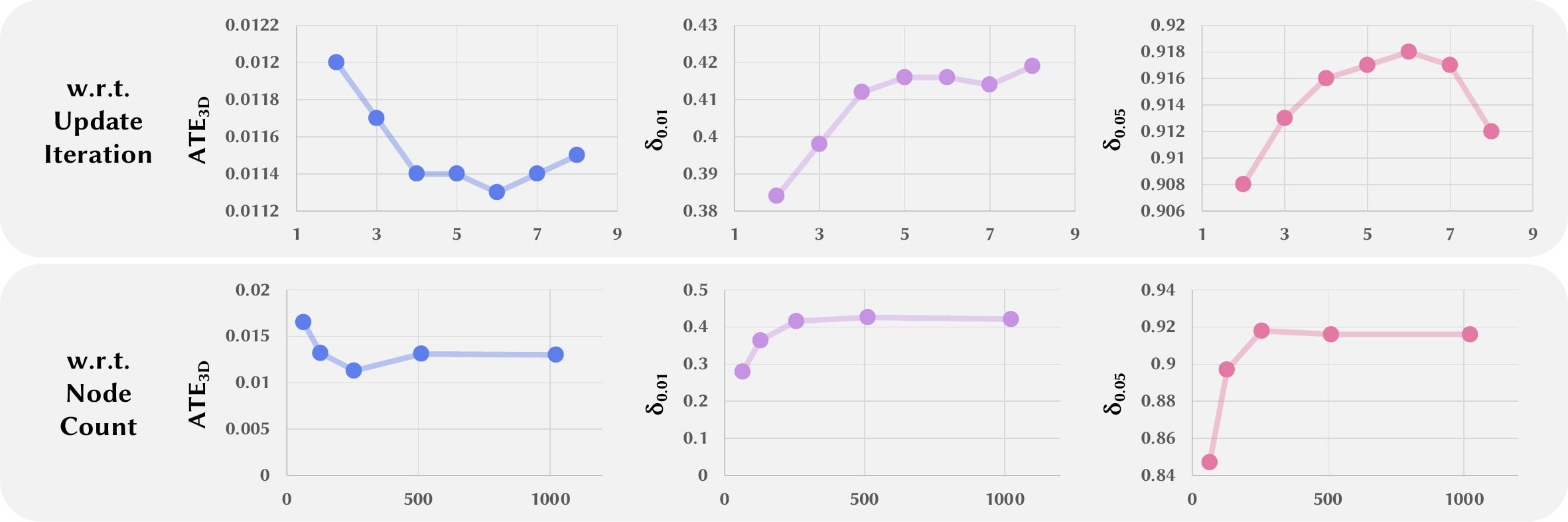}
    \vspace{-17pt}
    \caption{
        \textbf{Analysis on the effect of update iteration and node count.}
        Metrics $ATE_{3D}$, $\delta_{0.01}$ and $\delta_{0.05}$ w.r.t. transformer update iterations and node counts are presented.
        We update trajectories 6 times and set node count $B=256$ as a balance between performance and efficiency. 
    }
    \label{fig:update-iter}
    \vspace{-12pt}

    \label{fig:node-num}
\end{figure*}

\paragraph{Effectiveness of node graph.}
We show the effectiveness of our blending skinning approach by constructing a pipeline where we use our network to predict the temporal trajectory across observed point clouds for any point on the source object, which is named ``Dense matching''.
Since the memory consumption grows exponentially with node number, we iteratively pass all source points through the node graph predictor and aggregate the results.
The results on D-FAUST \cite{bogo2017dfaust} dataset in Tab.~\ref{tab:ablation} show our skinning approach not only significantly improves registration efficiency but also achieves higher accuracy compared with ``Dense matching''. 

\begin{table}
  \centering
  \caption{\textbf{Ablation studies.} 
  The results are evaluated on the D-FAUST test set. Average trajectory error $ATE_{3D}$, inlier proportions $\delta_{0.01}$ and $\delta_{0.05}$ are reported.
  The average registration time per frame $T_{avg}$ is reported for "Ours" and "Dense matching" for efficiency comparison.
  }
  \label{tab:ablation}
  \resizebox{0.48\textwidth}{!}{
  \begin{tabular}{lcccccccc}
    \toprule
    Method                          & $ATE_{3D}\downarrow$ & $\delta_{0.01}\uparrow$ & $\delta_{0.05}\uparrow$ & $T_{avg}\downarrow$ \\
    \midrule
    Ours                            & \textbf{0.011}       & \textbf{0.416}          & \textbf{0.918}          & 0.172     \\
    \midrule
    Dense matching & 0.014                & 0.299                   & 0.885                   & 4.645               \\
    \midrule
    Ours \textit{w/o} $\phi$        & 0.022                & 0.289                   & 0.845                   & \textbf{0.060}                   \\
    Ours \textit{w/o} $\Phi$        & 0.014                & 0.366                   & 0.889                   & 0.169                   \\
    \bottomrule
  \end{tabular}
  }
\end{table}

\paragraph{Effectiveness of coarse-to-fine scheme.}
One of our major insights is that performing sequential non-rigid registration with coarse-to-fine scheme allows higher accuracy and better generalizability.
To prove the effectiveness of our design, we construct two variants of our method by removing the node-to-frame matching network $\phi$ and sequential node regression network $\Phi$ respectively.
As is shown in Tab.~\ref{tab:ablation}, both variants show a significant drop in performance.

\paragraph{Analysis on update iteration.}
For our baseline method, we perform iterative updates with sequential transformer for $M=6$ iterations.
As comparison, we plot the evaluation results using different update iterations in Fig.~\ref{fig:update-iter}.
The results show that our model trained with $M=6$ performs best at $M=6$, but can be inferenced at $M=4$ for better efficiency with minor performance drop.

\paragraph{Analysis on node amount.}
\label{ablation:num-node}

Node amount significantly affects registration accuracy, since more nodes are capable of modeling more complex motion with more detail.
However, prediction large amount of graph nodes is computationally expensive.
We show the evaluation results for different node amount in Fig.~\ref{fig:node-num}, which shows our choice of $B=256$ nodes is the best balance between registration accuracy and efficiency.

\section{Conclusions}

In this work, we performs efficient sequential non-rigid registration by representing temporal correspondences as deformation graphs.
We designed a coarse-to-fine matching pipeline to estimate node trajectories of deformation graphs with strong generalization ability.
Based on the predicted node positions and radii, we additionally proposed an RBF-based LBS technique for deforming the source object to target point clouds.
Experiments demonstrate that our method outperforms existing methods both in accuracy and efficiency across public datasets.

\label{limitation}

Although our method displays great generalizability, it is trained on limited data due to the lack of annotated 4D correspondences, and could benefit from being trained on a larger variety of objects and motions.
Its further applications in other areas, such as dynamic scene editing, compression, autonomous driving and robotics, are still yet to be explored.

\paragraph{Acknowledgments.}
This work was partially supported by the National Key R\&D Program of China (No. 2024YFB2809102), NSFC (No. 624B1017, No. 62402427, No. U24B20154), Zhejiang Provincial Natural Science Foundation of China (No. LR25F020003), Zhejiang University Education Foundation Qizhen Scholar Foundation, and Information Technology Center and State Key Lab of CAD\&CG, Zhejiang University.

\newpage

{
    \small
    \bibliographystyle{ieeenat_fullname}
    \bibliography{ref}

\begin{thebibliography}{40}
\providecommand{\natexlab}[1]{#1}
\providecommand{\url}[1]{\texttt{#1}}
\expandafter\ifx\csname urlstyle\endcsname\relax
  \providecommand{\doi}[1]{doi: #1}\else
  \providecommand{\doi}{doi: \begingroup \urlstyle{rm}\Url}\fi

\bibitem[Allen et~al.(2003)Allen, Curless, and Popovi{\'c}]{allen2003space}
Brett Allen, Brian Curless, and Zoran Popovi{\'c}.
\newblock The space of human body shapes: reconstruction and parameterization from range scans.
\newblock \emph{ACM Trans. on Graphics (TOG)}, 22\penalty0 (3):\penalty0 587--594, 2003.

\bibitem[Bogo et~al.(2017)Bogo, Romero, Pons-Moll, and Black]{bogo2017dfaust}
Federica Bogo, Javier Romero, Gerard Pons-Moll, and Michael~J. Black.
\newblock Dynamic {FAUST}: {R}egistering human bodies in motion.
\newblock In \emph{IEEE Conference on Computer Vision and Pattern Recognition (CVPR)}, 2017.

\bibitem[Bozic et~al.(2021)Bozic, Palafox, Zollh{\"{o}}fer, Thies, Dai, and Nie{\ss}ner]{deformation_graph}
Aljaz Bozic, Pablo~R. Palafox, Michael Zollh{\"{o}}fer, Justus Thies, Angela Dai, and Matthias Nie{\ss}ner.
\newblock Neural deformation graphs for globally-consistent non-rigid reconstruction.
\newblock In \emph{IEEE Conference on Computer Vision and Pattern Recognition (CVPR)}, 2021.

\bibitem[Brown and Rusinkiewicz(2004)]{thinplate-splines}
Benedict~J. Brown and Szymon Rusinkiewicz.
\newblock Non-rigid range-scan alignment using thin-plate splines.
\newblock In \emph{International Symposium on 3D Data Processing, Visualization and Transmission}, pages 759--765. {IEEE} Computer Society, 2004.

\bibitem[Brown and Rusinkiewicz(2007)]{global-nonrigid-alignment}
Benedict~J. Brown and Szymon Rusinkiewicz.
\newblock Global non-rigid alignment of 3-d scans.
\newblock \emph{ACM Trans. on Graphics (TOG)}, 26\penalty0 (3):\penalty0 21–es, 2007.

\bibitem[Cao et~al.(2024)Cao, Luo, Zhang, Nie{\ss}ner, and Tang]{m2vs}
Wei Cao, Chang Luo, Biao Zhang, Matthias Nie{\ss}ner, and Jiapeng Tang.
\newblock Motion2vecsets: 4d latent vector set diffusion for non-rigid shape reconstruction and tracking.
\newblock \emph{CoRR}, abs/2401.06614, 2024.

\bibitem[Chang and Zwicker(2009)]{DBLP:journals/cgf/ChangZ09}
Will Chang and Matthias Zwicker.
\newblock Range scan registration using reduced deformable models.
\newblock \emph{Computer Graphics Forum}, 28\penalty0 (2):\penalty0 447--456, 2009.

\bibitem[Doersch et~al.(2022)Doersch, Gupta, Markeeva, Recasens, Smaira, Aytar, Carreira, Zisserman, and Yang]{doersch2022tap}
Carl Doersch, Ankush Gupta, Larisa Markeeva, Adria Recasens, Lucas Smaira, Yusuf Aytar, Joao Carreira, Andrew Zisserman, and Yi Yang.
\newblock Tap-vid: A benchmark for tracking any point in a video.
\newblock \emph{Advances in Neural Information Processing Systems (NeurIPS)}, 2022.

\bibitem[Doersch et~al.(2023)Doersch, Yang, Vecerik, Gokay, Gupta, Aytar, Carreira, and Zisserman]{doersch2023tapir}
Carl Doersch, Yi Yang, Mel Vecerik, Dilara Gokay, Ankush Gupta, Yusuf Aytar, Joao Carreira, and Andrew Zisserman.
\newblock Tapir: Tracking any point with per-frame initialization and temporal refinement.
\newblock In \emph{IEEE International Conference on Computer Vision (ICCV)}, 2023.

\bibitem[Huang et~al.(2011)Huang, Budd, and Hilton]{huang2011global}
Peng Huang, Chris Budd, and Adrian Hilton.
\newblock Global temporal registration of multiple non-rigid surface sequences.
\newblock In \emph{IEEE Conference on Computer Vision and Pattern Recognition (CVPR)}, 2011.

\bibitem[Huang et~al.(2021)Huang, Huang, Sun, Zhang, Jiang, and Bajaj]{huang2021arapreg}
Qixing Huang, Xiangru Huang, Bo Sun, Zaiwei Zhang, Junfeng Jiang, and Chandrajit Bajaj.
\newblock Arapreg: An as-rigid-as possible regularization loss for learning deformable shape generators.
\newblock In \emph{IEEE International Conference on Computer Vision (ICCV)}, 2021.

\bibitem[Jakab et~al.(2020)Jakab, Tucker, Makadia, Wu, Snavely, and Kanazawa]{jakab2021keypointdeformer}
Tomas Jakab, Richard Tucker, Ameesh Makadia, Jiajun Wu, Noah Snavely, and Angjoo Kanazawa.
\newblock Keypointdeformer: Unsupervised 3d keypoint discovery for shape control.
\newblock In \emph{IEEE Conference on Computer Vision and Pattern Recognition (CVPR)}, 2020.

\bibitem[Karaev et~al.(2023)Karaev, Rocco, Graham, Neverova, Vedaldi, and Rupprecht]{cotracker}
Nikita Karaev, Ignacio Rocco, Benjamin Graham, Natalia Neverova, Andrea Vedaldi, and Christian Rupprecht.
\newblock Cotracker: It is better to track together.
\newblock \emph{CoRR}, abs/2307.07635, 2023.

\bibitem[Lei and Daniilidis(2022)]{cadex}
Jiahui Lei and Kostas Daniilidis.
\newblock Cadex: Learning canonical deformation coordinate space for dynamic surface representation via neural homeomorphism.
\newblock In \emph{IEEE Conference on Computer Vision and Pattern Recognition (CVPR)}, 2022.

\bibitem[Li et~al.(2008{\natexlab{a}})Li, Sumner, and Pauly]{DBLP:journals/cgf/LiSP08}
Hao Li, Robert~W. Sumner, and Mark Pauly.
\newblock Global correspondence optimization for non-rigid registration of depth scans.
\newblock \emph{Computer Graphics Forum}, 27\penalty0 (5):\penalty0 1421--1430, 2008{\natexlab{a}}.

\bibitem[Li et~al.(2008{\natexlab{b}})Li, Sumner, and Pauly]{NICP}
Hao Li, Robert~W Sumner, and Mark Pauly.
\newblock Global correspondence optimization for non-rigid registration of depth scans.
\newblock In \emph{Computer Graphics Forum}, pages 1421--1430, 2008{\natexlab{b}}.

\bibitem[Li et~al.(2023)Li, Zheng, Ferroni, Pontes, and Lucey]{nsfp}
Xueqian Li, Jianqiao Zheng, Francesco Ferroni, Jhony~Kaesemodel Pontes, and Simon Lucey.
\newblock Fast neural scene flow.
\newblock In \emph{IEEE International Conference on Computer Vision (ICCV)}, 2023.

\bibitem[Li and Harada(2022{\natexlab{a}})]{lepard}
Yang Li and Tatsuya Harada.
\newblock Lepard: Learning partial point cloud matching in rigid and deformable scenes.
\newblock In \emph{IEEE Conference on Computer Vision and Pattern Recognition (CVPR)}, 2022{\natexlab{a}}.

\bibitem[Li and Harada(2022{\natexlab{b}})]{ndp}
Yang Li and Tatsuya Harada.
\newblock Non-rigid point cloud registration with neural deformation pyramid.
\newblock In \emph{Advances in Neural Information Processing Systems (NeurIPS)}, 2022{\natexlab{b}}.

\bibitem[Li et~al.(2021{\natexlab{a}})Li, Takehara, Taketomi, Zheng, and Nießner]{li20214dcomplete}
Yang Li, Hikari Takehara, Takafumi Taketomi, Bo Zheng, and Matthias Nießner.
\newblock 4dcomplete: Non-rigid motion estimation beyond the observable surface.
\newblock 2021{\natexlab{a}}.

\bibitem[Li et~al.(2021{\natexlab{b}})Li, Niklaus, Snavely, and Wang]{NSFF}
Zhengqi Li, Simon Niklaus, Noah Snavely, and Oliver Wang.
\newblock Neural scene flow fields for space-time view synthesis of dynamic scenes.
\newblock In \emph{IEEE Conference on Computer Vision and Pattern Recognition (CVPR)}, 2021{\natexlab{b}}.

\bibitem[Liao et~al.(2009)Liao, Zhang, Wang, Yang, and Gong]{pointwise_field}
Miao Liao, Qing Zhang, Huamin Wang, Ruigang Yang, and Minglun Gong.
\newblock Modeling deformable objects from a single depth camera.
\newblock In \emph{IEEE International Conference on Computer Vision (ICCV)}, 2009.

\bibitem[Liu et~al.(2019)Liu, Qi, and Guibas]{flownet3d}
Xingyu Liu, Charles~R Qi, and Leonidas~J Guibas.
\newblock Flownet3d: Learning scene flow in 3d point clouds.
\newblock \emph{IEEE Conference on Computer Vision and Pattern Recognition (CVPR)}, 2019.

\bibitem[Mescheder et~al.(2019)Mescheder, Oechsle, Niemeyer, Nowozin, and Geiger]{occnet}
Lars~M. Mescheder, Michael Oechsle, Michael Niemeyer, Sebastian Nowozin, and Andreas Geiger.
\newblock Occupancy networks: Learning 3d reconstruction in function space.
\newblock In \emph{IEEE Conference on Computer Vision and Pattern Recognition (CVPR)}, pages 4460--4470. Computer Vision Foundation / {IEEE}, 2019.

\bibitem[Mildenhall et~al.(2020)Mildenhall, Srinivasan, Tancik, Barron, Ramamoorthi, and Ng]{mildenhall2020nerf}
Ben Mildenhall, Pratul~P Srinivasan, Matthew Tancik, Jonathan~T Barron, Ravi Ramamoorthi, and Ren Ng.
\newblock Nerf: Representing scenes as neural radiance fields for view synthesis.
\newblock In \emph{European Conference on Computer Vision (ECCV)}, 2020.

\bibitem[Newcombe et~al.(2015)Newcombe, Fox, and Seitz]{DBLP:conf/cvpr/NewcombeFS15}
Richard~A. Newcombe, Dieter Fox, and Steven~M. Seitz.
\newblock Dynamicfusion: Reconstruction and tracking of non-rigid scenes in real-time.
\newblock In \emph{IEEE Conference on Computer Vision and Pattern Recognition (CVPR)}, 2015.

\bibitem[Niemeyer et~al.(2019)Niemeyer, Mescheder, Oechsle, and Geiger]{oflow}
Michael Niemeyer, Lars~M. Mescheder, Michael Oechsle, and Andreas Geiger.
\newblock Occupancy flow: 4d reconstruction by learning particle dynamics.
\newblock In \emph{IEEE International Conference on Computer Vision (ICCV)}, 2019.

\bibitem[Park et~al.(2021)Park, Sinha, Barron, Bouaziz, Goldman, Seitz, and Martin{-}Brualla]{Nerfies}
Keunhong Park, Utkarsh Sinha, Jonathan~T. Barron, Sofien Bouaziz, Dan~B. Goldman, Steven~M. Seitz, and Ricardo Martin{-}Brualla.
\newblock Nerfies: Deformable neural radiance fields.
\newblock In \emph{IEEE International Conference on Computer Vision (ICCV)}, 2021.

\bibitem[Qi et~al.(2017)Qi, Su, Mo, and Guibas]{pointnet}
Charles~Ruizhongtai Qi, Hao Su, Kaichun Mo, and Leonidas~J. Guibas.
\newblock Pointnet: Deep learning on point sets for 3d classification and segmentation.
\newblock \emph{IEEE Conference on Computer Vision and Pattern Recognition (CVPR)}, 2017.

\bibitem[Qin et~al.(2023{\natexlab{a}})Qin, Yu, Wang, Guo, Peng, Ilic, Hu, and Xu]{geotr}
Zheng Qin, Hao Yu, Changjian Wang, Yulan Guo, Yuxing Peng, Slobodan Ilic, Dewen Hu, and Kai Xu.
\newblock Geotransformer: Fast and robust point cloud registration with geometric transformer.
\newblock \emph{IEEE Trans. on Pattern Analysis and Machine Intelligence (PAMI)}, 45\penalty0 (8):\penalty0 9806--9821, 2023{\natexlab{a}}.

\bibitem[Qin et~al.(2023{\natexlab{b}})Qin, Yu, Wang, Peng, and Xu]{Deepgraph}
Zheng Qin, Hao Yu, Changjian Wang, Yuxing Peng, and Kai Xu.
\newblock Deep graph-based spatial consistency for robust non-rigid point cloud registration.
\newblock In \emph{IEEE Conference on Computer Vision and Pattern Recognition (CVPR)}, 2023{\natexlab{b}}.

\bibitem[Ronneberger et~al.(2015)Ronneberger, Fischer, and Brox]{unet}
Olaf Ronneberger, Philipp Fischer, and Thomas Brox.
\newblock U-net: Convolutional networks for biomedical image segmentation.
\newblock \emph{CoRR}, abs/1505.04597, 2015.

\bibitem[Sharf et~al.(2008)Sharf, Alcantara, Lewiner, Greif, Sheffer, Amenta, and Cohen{-}Or]{DBLP:journals/tog/SharfALGSAC08}
Andrei Sharf, Dan~A. Alcantara, Thomas Lewiner, Chen Greif, Alla Sheffer, Nina Amenta, and Daniel Cohen{-}Or.
\newblock Space-time surface reconstruction using incompressible flow.
\newblock \emph{ACM Trans. on Graphics (TOG)}, 27\penalty0 (5):\penalty0 110, 2008.

\bibitem[Sumner et~al.(2007{\natexlab{a}})Sumner, Schmid, and Pauly]{DBLP:journals/tog/SumnerSP07}
Robert~W. Sumner, Johannes Schmid, and Mark Pauly.
\newblock Embedded deformation for shape manipulation.
\newblock \emph{ACM Trans. on Graphics (TOG)}, 26\penalty0 (3):\penalty0 80, 2007{\natexlab{a}}.

\bibitem[Sumner et~al.(2007{\natexlab{b}})Sumner, Schmid, and Pauly]{deformGraph}
Robert~W Sumner, Johannes Schmid, and Mark Pauly.
\newblock Embedded deformation for shape manipulation.
\newblock In \emph{ACM Trans. on Graphics (TOG)}, pages 80--es. 2007{\natexlab{b}}.

\bibitem[Tang et~al.(2021)Tang, Xu, Jia, and Zhang]{lpdc}
Jiapeng Tang, Dan Xu, Kui Jia, and Lei Zhang.
\newblock Learning parallel dense correspondence from spatio-temporal descriptors for efficient and robust 4d reconstruction.
\newblock In \emph{IEEE Conference on Computer Vision and Pattern Recognition (CVPR)}, 2021.

\bibitem[Tevs et~al.(2012)Tevs, Berner, Wand, Ihrke, Bokeloh, Kerber, and Seidel]{DBLP:journals/tog/TevsBWIBKS12}
Art Tevs, Alexander Berner, Michael Wand, Ivo Ihrke, Martin Bokeloh, Jens Kerber, and Hans{-}Peter Seidel.
\newblock Animation cartography - intrinsic reconstruction of shape and motion.
\newblock \emph{ACM Trans. on Graphics (TOG)}, 31\penalty0 (2):\penalty0 12:1--12:15, 2012.

\bibitem[Wang et~al.(2019)Wang, Chen, Li, and Fang]{reg_network}
Lingjing Wang, Jianchun Chen, Xiang Li, and Yi Fang.
\newblock Non-rigid point set registration networks.
\newblock \emph{CoRR}, abs/1904.01428, 2019.

\bibitem[Xiao et~al.(2024)Xiao, Wang, Zhang, Xue, Peng, Shen, and Zhou]{spatialtracker}
Yuxi Xiao, Qianqian Wang, Shangzhan Zhang, Nan Xue, Sida Peng, Yujun Shen, and Xiaowei Zhou.
\newblock Spatialtracker: Tracking any 2d pixels in 3d space.
\newblock \emph{IEEE Conference on Computer Vision and Pattern Recognition (CVPR)}, abs/2404.04319, 2024.

\bibitem[Yu et~al.(2023)Yu, Qin, Hou, Saleh, Li, Busam, and Ilic]{roitr}
Hao Yu, Zheng Qin, Ji Hou, Mahdi Saleh, Dongsheng Li, Benjamin Busam, and Slobodan Ilic.
\newblock Rotation-invariant transformer for point cloud matching.
\newblock In \emph{IEEE Conference on Computer Vision and Pattern Recognition (CVPR)}, 2023.

\end{thebibliography}
}

\end{document}